\documentclass[letterpaper,journal]{IEEEtran}
\usepackage{amsmath,amsfonts}
\usepackage{algorithmic}
\usepackage{algorithm}
\usepackage{array}
\usepackage[caption=false,font=normalsize,labelfont=sf,textfont=sf]{subfig}
\usepackage{textcomp}
\usepackage{stfloats}
\usepackage{url}
\usepackage{verbatim}
\usepackage{graphicx}
\usepackage{xcolor}
\usepackage[authoryear,round, sort]{natbib}
\usepackage{booktabs}
\usepackage{enumitem}
\usepackage{hyperref}
\begin{document}

\title{Tactile-Guided Robot Learning: From Representation Learning to Foundation Models}

\author{Kiyanoush Nazari}
\author{Luchen Li}
\author{Haoshu Fang}
\author{Nima Fazeli}
\author{Alex Li}
\author{Thomas George Thuruthel}

\author{Kiyanoush Nazari*, Luchen Li, Haoshu Fang, Nima Fazeli, Zhibin Li, \\ Fumiya Iida, Thomas George Thuruthel 
\thanks{Kiyanoush Nazari (nkiyanoush@gmail.com), Luchen Li, Zhibin Li, and Thomas George Thuruthel are with the School of Computer Science, University College London. Haoshu Fang is with MIT CSAIL, Massachusetts Institute of Technology; Nima Fazeli is with the Department of Robotics, University of Michigan; and Fumiya Iida is with the School of Engineering, University of Tokyo.}
\thanks{Manuscript received October 13, 2026.}}



\maketitle

\begin{abstract}
Tactile sensing is becoming central to robust, adaptive, and dexterous robotic manipulation. Beyond its traditional role in contact detection and reactive control, touch is increasingly used to learn transferable representations, condition manipulation policies, and support embodied physical reasoning. Yet existing surveys largely emphasize tactile hardware and perception, leaving the rapid shift toward learning-based tactile manipulation insufficiently synthesized. This paper reviews tactile-guided skill learning from a learning-centric perspective, organized by the functional role of touch across the learning and control pipeline. We examine tactile representation and policy learning, tactile foundation models, predictive and adaptive control, and generalization across objects, tasks, sensors, embodiments, and simulation-to-real transfer. Across these directions, we identify common design principles and remaining challenges in learning transferable physical representations, integrating semantic reasoning with contact-level control, and scaling tactile learning across heterogeneous data and hardware. Finally, we outline emerging directions toward tactile foundation and world models that can predict, reason about, and adapt physical interactions. Together, this review characterizes the transition of touch from a task-specific feedback signal toward a core modality for general-purpose embodied intelligence. An accompanying database of the reviewed literature and associated resources is available at \href{https://sites.google.com/view/touch-in-robot-learning-review/home}{paper web page}.
\end{abstract}

\begin{IEEEkeywords}
Tactile sensing, robot learning, robotic manipulation, tactile world models.
\end{IEEEkeywords}





\section{Introduction}

Dexterous manipulation remains a major challenge in developing robots capable of human-level adaptability in unstructured environments \citep{lepora2024future}. Despite substantial advances in perception, planning, control, locomotion, and embodied AI, robots still struggle with everyday contact-rich tasks such as precision insertion, deformable-object manipulation, and assembly under uncertainty \citep{billard2019trends}. Unlike more structured robotic tasks, manipulation requires continuous regulation of physical interactions among the robot, objects, and environment \citep{cui2021toward}. Successful manipulation therefore depends not only on detecting contact, but on estimating and regulating evolving contact locations, forces, friction, compliance, and interaction dynamics, often across multiple simultaneous contacts. These quantities are difficult to infer reliably from vision alone, particularly under occlusion and during fine manipulation, making tactile perception and tactile-guided control essential for robust and dexterous interaction \citep{li2025developments}.

Recent tactile sensors can capture high-resolution contact geometry, force distributions, deformation, and slip \citep{genesis2026gene265, zhao2025embedding}, while advances in imitation learning \citep{tsuji2025survey}, reinforcement learning \citep{kim2026robot}, diffusion policies \citep{chi2025diffusion}, Vision-Language-Action (VLA) models \citep{ma2026survey}, and embodied AI \citep{feng2026multi} have substantially expanded robots' ability to learn complex manipulation skills from data \citep{o2024open}. Yet these developments remain only partially connected, where modern policies are still predominantly vision-centric and often treat touch as auxiliary information \citep{comanici2025gemini, black2024pi_0}. This limits their ability to reason about physical interactions in tasks where contact dynamics, rather than visual appearance alone, determine successful execution.

The role of touch in robot learning is consequently shifting from estimating low-level quantities such as contact, slip, force, and object properties \citep{howe1993tactile} toward providing physically grounded information throughout the learning and control pipeline. Tactile observations now support representation learning \citep{higuera2024sparsh}, multimodal policy learning, physical reasoning \citep{huang2025tactile, morissette2026tactile, yang2024binding}, adaptive decision making, and embodied world modeling \citep{cheng2026omnivtla}. This shift reframes the central question from \emph{how robots can sense touch} to \emph{how robots can learn from and act through touch}.

\begin{figure*}
    \centering
    \includegraphics[width=0.9\linewidth,trim={0 0 0.2cm 0},clip]{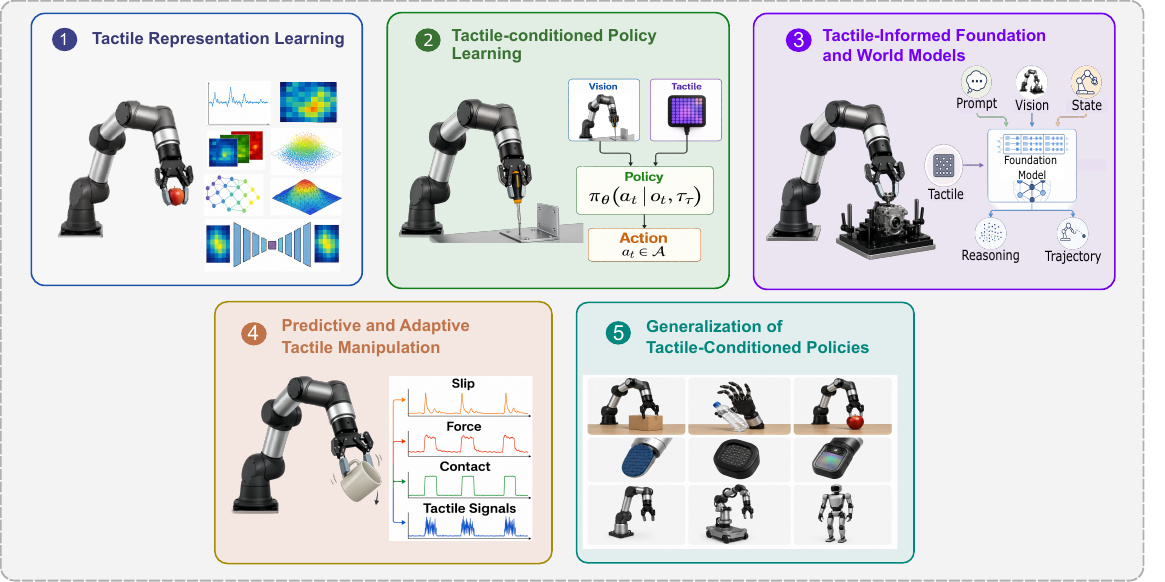}
    \caption{Scope of tactile-guided skill learning for robotic manipulation. We organize the literature according to the functional role of touch, spanning tactile representation learning, tactile-conditioned policies, tactile-informed foundation and world models, predictive and adaptive control, and generalization across objects, tasks, sensors, embodiments, environments, and the simulation-to-real gap.}
    \label{fig:scope}
\end{figure*}

Despite rapidly growing work across tactile sensing, robot learning, computer vision, and embodied AI, this literature remains fragmented. Existing surveys largely emphasize sensor hardware, sensing principles, classical tactile perception, or earlier tactile-guided manipulation and control \citep{li2020review, kroemer2021review}. Many also predate recent developments in foundation models and VLA systems and therefore do not capture the emerging role of touch within modern learning-based manipulation \citep{lepora2024future, jin2023progress, meribout2024tactile, mandil2023tactile, jahanshahi2025comprehensive, cho2024current, yu2025recent, zhang2022hardware, luo2025tactile, fang2025force, xie2025towards, hu2023machine, yao2024recent, xu2023recent, liu2023progress, huang2024recent, man2022recent, zhu2022recent, kroemer2021review}. A unified learning-centric account of how tactile information contributes across representation, policy learning, embodied reasoning, and generalization is therefore still lacking.

To address this gap, we organize the review according to the \emph{functional role of tactile information} within learning-based manipulation (Fig.~\ref{fig:scope}). We examine six research questions. \textbf{RQ1} considers how tactile observations are represented, encoded, and fused with other modalities, covering self-supervised \citep{dave2024multimodal, xu2025unit}, multimodal \citep{cheng2025touch100k, feng2026anytouch}, and cross-modal representation learning \citep{fu2024a}. \textbf{RQ2} examines how touch conditions action generation through imitation learning \citep{11455918}, reinforcement learning \citep{miller2026enhancing}, diffusion policies \citep{xue2025reactive}, and multimodal policy architectures. \textbf{RQ3} investigates tactile-informed foundation and world models, including tactile-enhanced VLAs \citep{hao2026tla}, multimodal world models \citep{zhang2026contactworld}, and foundation controllers \citep{you2026tebi}, where touch supports physically grounded reasoning and planning. \textbf{RQ4} considers predictive and adaptive tactile control, where tactile feedback supports interaction-state estimation, feedback regulation, contact prediction, online optimization, and adaptation under uncertain physical interactions \citep{yin2025learning,xu2024letac,zhao2026tacman,nazari2025bioinspired,pmlr-v205-nazari23a}. \textbf{RQ5} evaluates generalization beyond training conditions across objects, tasks, sensors, robot embodiments, environments, and simulation-to-real transfer. Finally, \textbf{RQ6} synthesizes these developments to identify the remaining challenges and research directions toward scalable tactile foundation models, general-purpose manipulation, and lifelong tactile learning.

\section{Tactile Representation Learning}
\label{sec:representation}


Tactile sensing provides localized observations of physical interaction shaped by contact dynamics, sensor mechanics, and robot motion \citep{johansson2009coding}. Raw tactile measurements, however, are often high-dimensional and sensor-dependent, with manipulation-relevant properties such as contact geometry, loading, deformation, and slip only implicitly encoded. Tactile representations bridge sensing and action by extracting compact, task-relevant features while preserving the spatial and temporal information required for perception, prediction, and control. While early approaches relied on physically motivated descriptors such as marker displacement, pressure distributions, and vibration spectra, recent work increasingly learns representations through supervised, self-supervised, multimodal, and multi-sensor training \citep{cao2026tactile}, aiming for reusable features with reduced dependence on task-specific labels and sensor characteristics.

This section examines four aspects of tactile representation learning, including learning informative features from unlabeled interaction data (Sec.~III-A); fusing touch with complementary sensing modalities (Sec.~III-B); aligning touch with pre-trained vision-language embedding spaces to exploit their semantic structure (Sec.~III-C); and, finally, assessing whether representations learned on particular sensors, tasks, or embodiments transfer beyond their training conditions (Sec.~III-D).

\subsection{Self-Supervised Tactile Representation Learning}

Self-supervised learning enables tactile representations to be learned from unlabeled interaction data \citep{gui2024survey, lee2018making}, avoiding physically motivated preprocessing such as marker tracking \citep{li2023marker}, finite-element modeling \citep{sferrazza2019ground}, or task-specific descriptors. Unlike vision, where large-scale datasets and pretrained models are readily available \citep{russakovsky2015imagenet}, tactile data require physical interaction and remain strongly sensor-dependent. Recent approaches therefore exploit data collected through robotic exploration, demonstration, or in-the-wild interaction \citep{yang2022touch}, using reconstruction, contrastive learning, self-distillation, masked modeling, or action-conditioned prediction \citep{mandil2022action} to learn reusable tactile features.

Reconstruction-based methods require an encoded latent state to retain sufficient information for recovering the original observation, or components removed from it. Given a tactile observation $\mathbf{x}$, an encoder $f_{\theta}$ produces $\mathbf{z}=f_{\theta}(\mathbf{x})$, from which a decoder $g_{\phi}$ reconstructs $\hat{\mathbf{x}}=g_{\phi}(\mathbf{z})$:

\begin{equation}
\mathcal{L}_{\mathrm{rec}}
=
\mathbb{E}{\mathbf{x}\sim p(\mathbf{x})}
\left[
d\left(
\mathbf{x},
g_{\phi}(f_{\theta}(\mathbf{x}))
\right)
\right],
\end{equation}
where $d(\cdot,\cdot)$ measures reconstruction error. Although scalable, this objective may preserve sensor-specific appearance rather than physically relevant interaction variables. Early work \citep{mura2020grasp} used a recurrent conditional autoencoder to encode fingertip-force sequences for vision-free grasping, transferring the representation to material recognition and grasp-stability estimation. UniT \citep{xu2025unit} instead introduces vector quantization \citep{van2017neural}, using a VQGAN to encode tactile images as discrete tokens; despite pretraining on a single simple object, its structured latent space transfers zero-shot to unseen-object pose estimation and classification and to three real-world policy-learning tasks. Following masked modeling \citep{he2022masked}, TacMAE \citep{cao2023mask} reconstructs masked tactile-image regions, achieving 71.4\% zero-shot and 85.8\% fine-tuned accuracy on unseen YCB objects. For structured 3D tactile arrays, \citep{wu2025canonical} reconstructs masked local forces from taxel coordinates, poses, and forces before transferring the encoder to a visuo-tactile Diffusion Policy. Finally, \citep{li2025tactile} represents tactile observations as modulated neural implicit functions using a shared SIREN trunk, substantially reducing storage while retaining information for downstream pose estimation. Collectively, these methods progress from reconstructing raw sequences and pixels toward discrete tokens, physical force fields, and continuous functions, introducing stronger structural priors to encourage more transferable representations.

\begin{figure*}
    \centering
    \includegraphics[width=\linewidth,trim={0 0 0cm 0},clip]{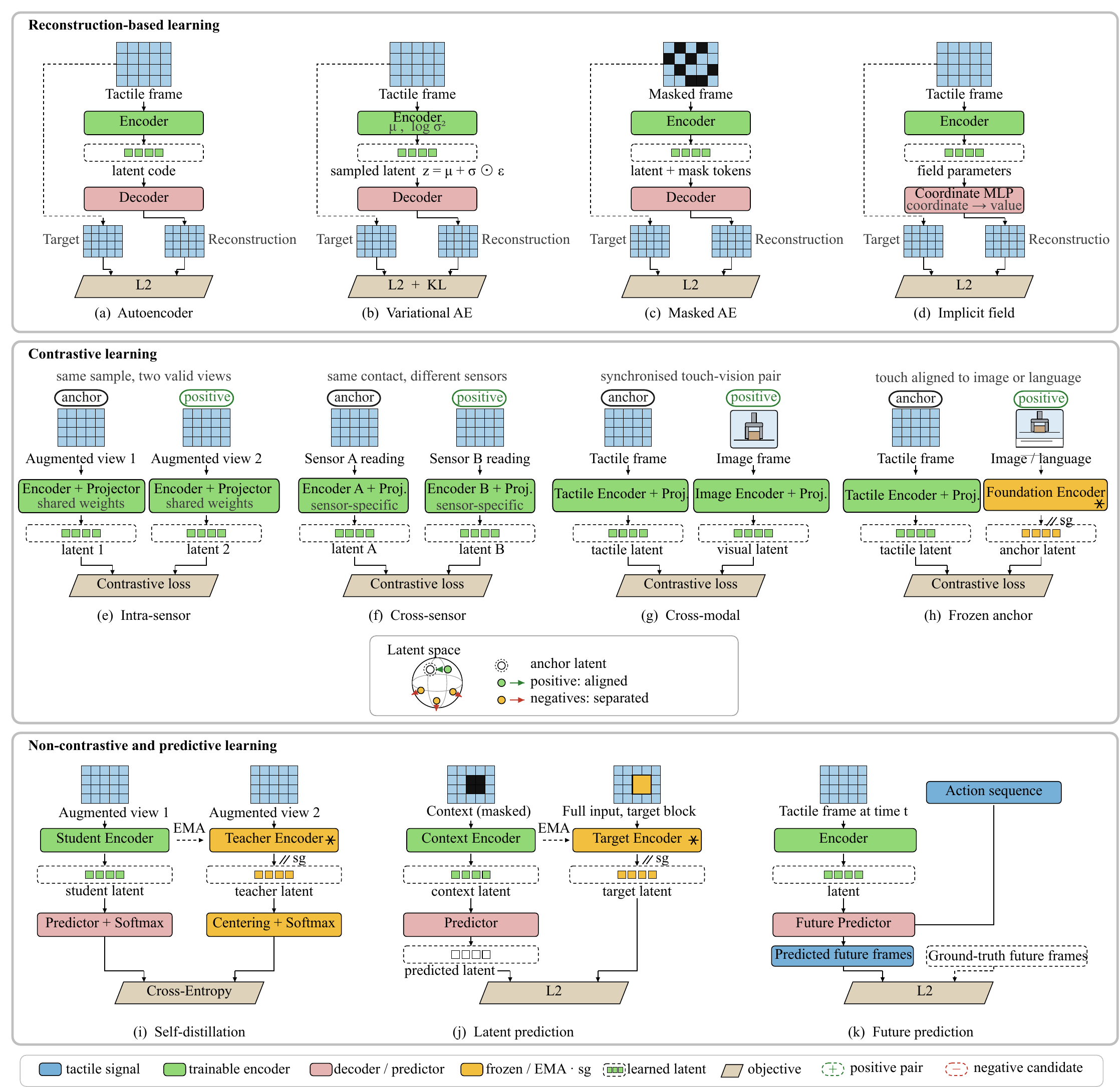}
    \caption{Self-supervised representation learning paradigms for tactile sensing. Reconstruction-based methods (a-d) recover a masked or transformed observation; contrastive methods (e-h) pull positive pairs together in a shared latent space while separating negatives; non-contrastive and predictive methods (i-k) match or predict latent targets without explicit negatives.}
    \label{fig:ssl}
\end{figure*}

Contrastive learning methods forgo reconstruction and instead enforce similarity between observations of the same interaction while separating distinct interactions. Given an anchor embedding $\mathbf{z}_i$, the following InfoNCE-style \citep{oord2018representation} objective increases its similarity to a positive embedding $\mathbf{z}_i^{+}$ while reducing its similarity to negative embeddings $\mathbf{z}_j^{-}$:
\begin{equation}
-\log
\frac{
\exp\left(
\operatorname{sim}(\mathbf{z}_i,\mathbf{z}_i^{+})/\tau
\right)
}{
\exp\left(
\operatorname{sim}(\mathbf{z}_i,\mathbf{z}_i^{+})/\tau
\right)
+
\sum_{j}
\exp\left(
\operatorname{sim}(\mathbf{z}_i,\mathbf{z}_j^{-})/\tau
\right)
},
\end{equation}
where $\operatorname{sim}(\cdot,\cdot)$ denotes similarity and $\tau$ is a temperature parameter. The four settings in Fig.~\ref{fig:ssl}e-h differ primarily in how positive pairs are defined, augmented views of the same reading, cross-sensor observations of the same contact, synchronized touch-vision pairs, or alignment to a frozen foundation-model anchor. MViTac \citep{dave2024multimodal} combines intra-modal contrast between augmented tactile views with inter-modal touch-vision alignment, outperforming either objective alone on material classification and grasp-success prediction. Cross-sensor pairs similarly underpin Contrastive Touch-to-Touch Pretraining \citep{rodriguez2025contrastive}, discussed in Section~III-D, while foundation model and synchronized multimodal alignment are explored by UniTouch \citep{yang2024binding}, ViTaL \citep{george2024vital}, and VITaMIn \citep{liu2025vitamin} in Section~III-C. Thus, the choice of positive pair directly determines which variations the representation treats as invariant.

Non-contrastive methods replace explicit negatives with consistency or prediction objectives (see Fig.~\ref{fig:ssl} for architecture examples). BYOL- \citep{grill2020bootstrap} and DINO-style \citep{caron2021emerging} self-distillation align online and target networks across augmented views, while spatial and temporal predictive methods infer masked or future latent states from context. Their central challenge remains defining appropriate invariances, since variations in contact location or pressure may represent either nuisance or task-relevant physical information. T-DEX \citep{guzey2023dexterity} uses BYOL to pretrain a tactile encoder on 2.5 hours of teleoperated play with distributed uSkin sensors, enabling nearest-neighbor action retrieval from a visuo-tactile latent space and data-efficient manipulation from few demonstrations. exUMI \citep{xu2025exumi} instead introduces Tactile Predictive Pretraining, using a Transformer with latent diffusion to predict future tactile states conditioned on context and actions; the frozen encoder is subsequently transferred to Diffusion Policies, preserving deformation and contact-transition information overlooked by static objectives.

Sparsh \citep{higuera2024sparsh} provides a large-scale comparison of these paradigms, benchmarking masked autoencoding, DINO self-distillation \citep{caron2021emerging}, and I-JEPA/V-JEPA \citep{assran2023self} on more than 460,000 unlabeled tactile observations through a benchmark of six touch-centric tasks. The stronger transfer of DINO and I-JEPA indicates that reconstructing tactile appearance does not necessarily preserve the mechanical information required downstream. More broadly, reconstruction, contrastive, and predictive objectives differ primarily in what they define as physically relevant structure: force fields, cross-modal semantics, augmentation invariance, or action-conditioned contact evolution. Their effectiveness therefore depends not simply on scale, but on whether the learning objective preserves interaction variables relevant to downstream manipulation. These methods nevertheless primarily learn representations from individual tactile streams; we next consider representations learned jointly with complementary sensory modalities.

\subsection{Multimodal Touch Representations}

Multimodal representation learning exploits complementary information across sensing modalities, analogous to human perception where visual and haptic cues are integrated according to their reliability \citep{ernst2002humans}. Vision provides global context but degrades under contact-induced occlusion, whereas touch directly captures local interaction states. Their fusion can be expressed as
\begin{equation}
z_t = F_{\phi}\!\left(
f_{I}\!\left(I_t\right),\,
f_{T}\!\left(T_t\right),\,
f_{p}\!\left(p_t\right)
\right),
\label{eq:fusion}
\end{equation}
where $f_I$, $f_T$, and $f_p$ are vision, tactile, and proprioception modality-specific encoders and $F_{\phi}$ is a fusion operator, producing the representation consumed by the policies in Section~IV. Fusion ranges from concatenating raw or independently encoded features to transformer-based joint token processing. M3L \citep{sferrazza2023power} instead learns fusion through joint masked reconstruction, training a Vision Transformer to reconstruct masked visual and tactile patches through cross-modal attention. The learned representation improves sample efficiency in downstream reinforcement learning and can benefit vision-only control even when touch is unavailable at deployment.

The simplest approaches concatenate independently encoded modalities, as in the ACT-based framework of \citep{zhao2026vitac} for deformable-object tracing and the contact-phase VITAL controller of \citep{zhao2025touch}. While being modular, concatenation leaves cross-modal correspondence to be learned implicitly. 3D-ViTac \citep{huang3d} instead projects vision and touch into a common 3D frame, forming a unified visuo-tactile point cloud in which contact locations and forces are directly grounded in scene geometry; Section~IV revisits this representation for diffusion-policy control.

Attention-based fusion learns cross-modal relationships without explicitly specifying geometry. ViTacFormer \citep{heng2025vitacformer} uses bidirectional cross-attention between visual and tactile features, together with future tactile-token prediction, improving bimanual manipulation under occlusion and sustained contact. Because high-bandwidth vision can otherwise dominate learning, \citep{li2025adaptive} introduces force-guided attention with auxiliary future-force prediction, achieving 93\% success across three contact-rich tasks while dynamically shifting attention toward touch during contact and vision during free motion. VHTformer \citep{li2025vhtformer} instead uses symmetric joint queries across vision, touch, and language to avoid structurally privileging any modality.

Multimodal tactile representations need not include vision. Sparsh-X \citep{higuera2025tactile} jointly models optical deformation, contact-microphone vibration, motion, and pressure from Digit 360, outperforming tactile-image-only representations on plug insertion and in-hand rotation. Overall, multimodal representations span simple feature concatenation, geometrically grounded fusion, learned cross-attention, and adaptive modality weighting, with their key distinction being how explicitly they encode correspondence and complementarity among sensory signals.

\subsection{Cross-Modal Alignment and Grounding}

Large pre-trained vision-language models provide shared embedding spaces into which additional modalities can be aligned \citep{girdhar2023imagebind}, an approach increasingly extended to touch \citep{yang2024binding, george2024vital, fu2024touch}. Unlike multimodal fusion (Sec.~III-B), which combines simultaneous observations into a task-specific representation, alignment projects touch into a pre-trained semantic space. Given paired data $(T,x)$, a tactile encoder $f_\phi$ is trained to match the embedding $g(x)$ of a frozen encoder, typically through an InfoNCE objective
\begin{equation}
\mathcal{L}_{\mathrm{align}}
= -\,\mathbb{E}
\left[
\log
\frac{\exp\!\left(\mathrm{sim}\!\left(f_{\phi}(T),\, g(x)\right)/\tau\right)}
{\sum_{j=1}^{N}\exp\!\left(\mathrm{sim}\!\left(f_{\phi}(T),\, g(x_j)\right)/\tau\right)}
\right],
\label{eq:align}
\end{equation}
following CLIP-style training \citep{radford2021learning}. This can transfer zero-shot semantic structure from large pretrained models \citep{yang2024binding, yu2025demonstrating}, but semantic alignment may discard manipulation-relevant quantities such as force and slip, motivating more physically grounded representations.

Synchronized vision-touch pairs provide the most direct alignment signal \citep{yang2022touch}. VITaL \citep{george2024vital} aligns visual and tactile observations so that touch shapes a visual representation that remains beneficial even when tactile input is removed at deployment. VITaMIn \citep{liu2025vitamin} introduces temporal structure by aligning a masked current observation with the CLIP embedding of a future unmasked frame, thereby relating contact to subsequent outcomes. Tac-DINO \citep{li2026tacdino} instead addresses spatial granularity: rather than aligning local touch with an entire image, it aligns touch with the corresponding visual patch, outperforming image-level alignment on over 20,000 contacts across 505 objects. Together, these approaches show that both the temporal target and spatial granularity of alignment determine what interaction structure is preserved.

UniTouch \citep{yang2024binding} aligns touch with frozen image embeddings already connected to language and sound, indirectly transferring this broader semantic structure and enabling zero-shot material recognition, cross-modal retrieval, and touch-conditioned generation. Octopi \citep{yu2024octopi} instead maps tactile embeddings into tokens consumed by a frozen large language model, enabling open-vocabulary reasoning about material and physical properties without explicit tactile-text alignment, although its reasoning remains dependent on the backbone's physical knowledge. Finally, FG-CLTP \citep{11599472} complements semantic descriptors with quantitative physical information by converting tactile deformation into 3D point clouds, discretizing force and contact geometry into numerical tokens, and jointly aligning them with point clouds and language. Its sensor-independent 3D representation additionally supports sim-to-real and cross-sensor transfer. Overall, these methods progressively extend tactile alignment from global semantic correspondence toward spatially localized, temporally predictive, multimodal, and physically grounded representations.

\subsection{Transferability and Generalization}

Recent work increasingly targets general-purpose tactile encoders rather than task-specific features \citep{higuera2024sparsh, zhao2024transferable, yang2024binding, feng2025anytouch}, following the pre-training paradigm of vision and language, 
$z = f_{\theta}(T),$ where $T \sim p_{\mathrm{src}}(T)$.
Such pre-training is valuable only if representations remain informative under shifts from $p_{\mathrm{src}}(T)$ to a target distribution $p_{\mathrm{tgt}}(T)$. This is particularly challenging for touch because measurements are tightly coupled to sensor hardware, with transfer failures occurring even between instances of the same sensor model due to manufacturing variation \citep{gupta2025sensor, agarwal2025modularized}.

Cross-sensor transfer follows three main strategies. First, joint multi-sensor pre-training seeks sensor-invariant structure from heterogeneous data. T3 \citep{zhao2024transferable} combines sensor-specific input encoders with a shared Transformer trained across 13 sensors and 11 tasks, while AnyTouch \citep{feng2025anytouch} learns cross-sensor correspondence in dynamic interactions and ToucHD \citep{feng2026anytouch} extends this toward hierarchical contact dynamics. These approaches produce reusable encoders but require large jointly curated multi-sensor datasets. Second, pairwise transfer learns mappings between particular sensors, GenForce \citep{chen2026training} translates heterogeneous observations into a shared marker representation while preserving source force labels; CTTP \citep{rodriguez2025contrastive} aligns GelSlim and Soft Bubble through InfoNCE; and Touch2Touch \citep{11127980} translates between them using diffusion. Such methods avoid large joint corpora but must be retrained for new sensor pairs. Third, SITR \citep{gupta2025sensor} explicitly conditions a modified Vision Transformer on calibration images, disentangling sensor-specific effects from contact geometry and thereby extending transfer to new sensors given a small calibration set.

These strategies trade data scale, transfer specificity, and deployment assumptions, where joint pre-training scales across sensors but requires heterogeneous corpora; pairwise methods reduce this requirement but scale poorly with sensor diversity; and disentanglement offers broader transfer at the cost of calibration. More importantly, transfer across tasks, objects, and robot embodiments remains comparatively underexplored at the representation level, and whether sensor-invariant features ultimately yield transferable manipulation policies remains an open question revisited in Section~VII.

Across the paradigms reviewed above, no tactile representation is universally optimal. Representations grounded in physical quantities such as force, contact geometry, or action-conditioned dynamics can improve invariance by abstracting away sensor-specific measurements, but excessive abstraction may discard fine texture, deformation, slip, or other task-relevant cues. The central challenge is therefore to preserve information required for downstream prediction and control while suppressing nuisance variation. Evaluating this balance remains difficult because benchmarks are fragmented across sensors, tasks, and datasets, motivating standardized evaluation of representation quality, transfer, and downstream manipulation performance. The following section examines how these representations are used for action generation.



\section{Tactile-Conditioned Policy Learning}
\label{sec:policy_learning}

Given an informative tactile representation, the next challenge is to exploit touch for action generation. Unlike classical tactile control \citep{howe1993tactile}, which typically reduces touch to interpretable variables for hand-crafted controllers, tactile-conditioned policies learn directly from rich tactile observations, allowing contact state, force distribution, slip, and compliance to influence actions as physical interactions evolve.

Formally, a policy receives multimodal observations
\begin{equation}
\label{eq}
o_t = \{I_t,T_t,p_t\},
\end{equation}
comprising visual, tactile, and proprioceptive signals. These may be encoded and fused into $z_t=f_{\phi}(o_t)$, from which the policy generates
\begin{equation}
a_t = \pi_{\theta}(z_t).
\end{equation}
While this observation-to-action formulation is broadly shared, learning paradigms differ in how $\pi_{\theta}$ is optimized and how tactile information is represented, fused, and exploited. We examine these choices across imitation learning, reinforcement learning, and diffusion-based policies. Methods that instead integrate learned tactile representations or dynamics within explicit feedback, predictive optimization, or adaptive control architectures are considered separately in Sec.~\ref{sec:adaptive_manipulation}.

\subsection{Imitation Learning for Contact-Rich Manipulation}

Imitation learning (IL) \citep{osa2018algorithmic} learns tactile-conditioned policies directly from expert demonstrations
\begin{equation}
\mathcal{D}=\{(o_t,a_t^*)\}_{t=1}^{N},
\end{equation}
typically through behavioral cloning (BC), where $o_t$ is the multimodal observation and $a_t^*$ the expert action \citep{jang2022bc}. The BC learning objectice is defined as,
\begin{equation}
\mathcal{L}_{\mathrm{BC}}
=
-\sum_{(o_t,a_t^*)\in\mathcal{D}}
\log \pi_{\theta}(a_t^*|o_t),
\label{eq:il_loss}
\end{equation}
Although the BC objective is largely shared, performance depends critically on how touch is represented, fused, and reflected in the demonstrations, since small changes in contact geometry can require substantially different actions.

\noindent  Because task demonstrations are expensive relative to unlabeled interaction, recent approaches increasingly pretrain tactile representations before imitation. T-DEX \citep{guzey2023dexterity} learns visual and tactile encoders self-supervised from robotic play, enabling dexterous skills from few demonstrations. exUMI \citep{xu2025exumi} instead uses Tactile Predictive Pretraining to predict future tactile states conditioned on observations and actions before transferring the encoder downstream. This distinction is important for IL, where static objectives primarily encode contact appearance, whereas action-conditioned prediction can capture how contact evolves under manipulation and thus provide more control-relevant features.

\noindent  Tactile, visual, and proprioceptive features are commonly encoded independently and concatenated before action prediction, as in ACT-based \citep{zhao2023learning} deformable-object tracing \citep{zhao2026vitac} and VITAL's local controller \citep{zhao2025touch}. Such fusion is modular but leaves spatial correspondence implicit. 3D-ViTac \citep{huang3d} instead projects vision and touch into a common 3D frame, representing them as a unified visuo-tactile point cloud that grounds contact locations and forces relative to scene geometry. These approaches illustrate a progression from treating touch as an additional feature vector toward explicitly encoding \emph{where} interaction occurs.

\noindent Scaling IL requires reducing dependence on task-specific robot teleoperation. T-DEX exploits task-agnostic robotic play \citep{guzey2023dexterity}, while exUMI \citep{xu2025exumi} uses a handheld visuo-tactile interface to collect action-aligned human interactions. DexVLA \citep{wen2025dexvla} combines VR teleoperation for macro arm motion with autonomous DexGrasp-VLA control using visual and tactile feedback for fine hand manipulation, reducing operator burden while collecting coordinated arm-hand demonstrations \citep{cui2025end}. To address embodiment mismatch, TactAlign \citep{wi2025tactalign} separately encodes human and robot tactile observations and uses rectified flow \citep{liu2022flow} to align their latent spaces from unpaired demonstrations, which subsequently condition a shared ACT policy \citep{zhao2023learning}. Together, these approaches broaden tactile demonstrations through play, portable human data collection, shared autonomy, and cross-embodiment alignment.

\noindent Tactile IL increasingly incorporates temporal and task structure beyond one-step BC. ACT \citep{zhao2023learning} predicts coherent action chunks conditioned on touch \citep{wi2025tactalign,zhao2026vitac}, while diffusion-based policies capture multimodal contact-rich action distributions \citep{xu2025exumi,huang3d}. Main components of tactile IL methods are included in Table~\ref{tab:tactile_policy_learning}. The tracing framework of \citep{zhao2026vitac} further combines BC with local contact-maintenance and global task-progress objectives, whereas VITAL \citep{zhao2025touch} separates coarse visual reaching from local visuo-tactile manipulation and augments the latter with RL-based residual corrections. These methods consistently place touch at the local interaction level, where continuous correction of contact and geometric errors is most critical.

Overall, tactile IL is progressing from task-specific BC toward reusable representations, spatially grounded fusion, and scalable demonstration sources. Vision primarily provides global task context, while touch resolves local physical states under contact and occlusion. However, IL remains constrained by its demonstration distribution, where tactile feedback may reveal an unfamiliar contact state without providing a recovery behavior absent from the training data. This motivates interactive policy improvement through tactile-conditioned reinforcement learning.

\begin{table*}[t]
\centering
\caption{Comparison of the major learning paradigms for tactile-conditioned policy learning.}
\label{tab:tactile_policy_learning}
\renewcommand{\arraystretch}{1.25}
\footnotesize
\begin{tabular}{p{2.2cm}p{2.4cm}p{2.8cm}p{2.6cm}p{2.8cm}p{2.8cm}}
\toprule
\textbf{Learning Paradigm}
&
\textbf{Optimization Objective}
&
\textbf{Tactile Integration}
&
\textbf{Typical Architectures}
&
\textbf{Typical Strengths}
&
\textbf{Main Limitations}
\\
\midrule

Imitation Learning

\hspace{-5pt}
\citep{guzey2023dexterity}
\citep{xu2025exumi}

\hspace{-5pt}
\citep{huang3d}
\citep{zhao2026vitac}

\hspace{-5pt}
\citep{wi2025tactalign}
\citep{zhao2025touch}
&
Behavior cloning from expert demonstrations
&
\vspace{-7pt}
\begin{itemize}[leftmargin=*, nosep, label=-]
    \item Feature fusion
    \item Pretrained tactile representations
    \item Latent conditioning
\end{itemize}
&
\vspace{-7pt}
\begin{itemize}[leftmargin=*, nosep, label=-]
\item Behavior Cloning
\item ACT
\item Transformer
\item Diffusion Policy
\item Nearest-Neighbor Retrieval
\end{itemize}
&
Data-efficient learning, stable training, precise contact behaviors
&
Limited to demonstrated behaviors, weak recovery from unseen failures
\\
\midrule

Reinforcement Learning

\hspace{-5pt}
\citep{yin2025learning}
\citep{tang2025visual}

\hspace{-5pt}
\citep{ye2026visual}
\citep{fieldtext2touch}

\hspace{-5pt}
\citep{church2022tactile}
&
Reward maximization through interaction
&
\vspace{-7pt}
\begin{itemize}[leftmargin=*, nosep, label=-]
    \item State representation
    \item Reward shaping
    \item Teacher-student learning
    \item Latent representations
\end{itemize}
&
\vspace{-7pt}
\begin{itemize}[leftmargin=*, nosep, label=-]
\item SAC
\item PPO
\item Actor-Critic
\item Teacher-Student Policies
\end{itemize}
&
Adaptive online interaction, robust recovery, force regulation
&
Sample inefficiency, reward engineering, sim-to-real challenges
\\
\midrule

Diffusion 

Policies

\hspace{-5pt}
\citep{xue2025reactive}
\citep{yin2025dexteritygen}

\hspace{-5pt}
\citep{wu2025tacdiffusion}
\citep{lin2025pp}

\hspace{-5pt}
\citep{zhu2026touch}
&
Conditional generative action modeling
&
\vspace{-5pt}
\begin{itemize}[leftmargin=*, nosep, label=-]
\item Observation conditioning
\item Latent conditioning
\item Reactive refinement
\end{itemize}
&
\vspace{-7pt}
\begin{itemize}[leftmargin=*, nosep, label=-]
\item DDPM
\item Latent Diffusion
\item Diffusion Policy
\item U-Net
\item Transformer
\end{itemize}
&
Multi-modal action generation, smooth trajectories, uncertainty modeling
&
Large demonstration datasets, iterative sampling, high computational cost
\\
\midrule

Visuo-Tactile Fusion

&
Joint multimodal policy learning
&
\vspace{-5pt}
\begin{itemize}[leftmargin=*, nosep, label=-]
\item Early/late fusion
\item cross-attention
\item token fusion
\item geometric fusion
\item adaptive gating

\end{itemize}
&
\vspace{-7pt}
\begin{itemize}[leftmargin=*, nosep, label=-]
\item PointNet
\item Cross-Attention
\item Token Fusion
\item Gated Fusion
\item Multimodal Transformers
\end{itemize}
&
Robust perception under occlusion and contact uncertainty
&
Fusion strategy remains task-dependent and architecture-specific
\\

\bottomrule
\end{tabular}
\end{table*}

\subsection{Reinforcement Learning with Tactile Feedback}

Reinforcement learning (RL) \citep{deisenroth2013survey} complements imitation learning by optimizing manipulation policies through interaction rather than demonstrated actions. Tactile-conditioned RL can be formulated as a Markov Decision Process
\begin{equation}
\mathcal{M}=(\mathcal{S},\mathcal{A},P,r,\gamma),
\end{equation}
where $\mathcal{S}$ and $\mathcal{A}$ are state and action spaces, $P$ the transition dynamics, $r$ the reward, and $\gamma$ the discount factor. Because the physical state is only partially observable, the policy operates on multimodal observations $o_t=\{I_t,T_t,p_t\}$ and optimizes
\begin{equation}
J(\theta)=
\mathbb{E}_{\pi_\theta}
\left[
\sum_{t=0}^{\infty}
\gamma^t r(s_t,a_t)
\right].
\end{equation}
Unlike behavioral cloning, RL must discover actions that maximize long-horizon outcomes, making it particularly suited to manipulation requiring active establishment and regulation of contact.

\noindent Touch reduces uncertainty about local physical states that remain latent from vision and proprioception. For peg insertion, \citep{tang2025visual} combines force/torque sensing with a self-supervised visual-proprioceptive representation and SAC \citep{haarnoja2018soft} to predict Cartesian corrections under pose uncertainty, executed through admittance control. For in-hand translation, \citep{yin2025learning} shows that normal and shear tactile signals improve PPO \citep{schulman2017proximal} robustness to unseen geometries, center-of-mass variations, and hand orientations. Because learning such states directly from high-dimensional observations increases RL sample complexity \citep{zhang2025settling}, representations are often pretrained separately, \citep{tang2025visual} freezes a predictive multimodal encoder before SAC training, while \citep{ye2026visual} pretrains a masked-autoencoder Transformer \citep{he2022masked} on visual and binary tactile demonstrations before combining its features with proprioception and task information. The latter further distills task-specific simulated PPO \citep{schulman2017proximal} experts into a unified policy through online imitation. These approaches separate \emph{learning the contact state} from \emph{learning how to act upon it}.

\noindent Tactile RL faces a distinctive exploration challenge, where informative tactile observations arise only through physical interaction, yet unrestricted exploration can produce excessive forces, unstable contacts, or hardware damage. Structured strategies, such as reward-guided random exploration in \citep{tang2025visual}, therefore constrain data collection, while tactile quantities themselves can shape rewards through force regulation, contact stability, or insertion quality. Designing such objectives becomes difficult when contact goals are multidimensional. Text2Touch \citep{fieldtext2touch} uses large language models to iteratively generate and refine reward functions for PPO-based dexterous manipulation, illustrating that richer tactile observations must be accompanied by effective mechanisms for converting physical interaction variables into learning signals.

\noindent To reduce costly real-world exploration, tactile RL increasingly exploits simulation and privileged teacher-student learning. \citep{yin2025learning} trains a PPO teacher using privileged object pose, velocity, mass, center of mass, and friction, then distills its latent representation and actions into a Transformer student conditioned only on proprioception and simulated touch. Text2Touch \citep{fieldtext2touch} similarly transfers behavior from privileged simulation to a tactile-proprioceptive student. This formulation is particularly suited to tactile manipulation, simulation exposes physical variables unavailable at deployment, while touch provides an observable interface through which the student can infer their action-relevant effects, including changes in friction, inertia, contact geometry, and disturbances.

\noindent Simulation improves RL scalability and safety but introduces a substantial tactile domain gap because sensor observations depend on deformation, contact mechanics, material properties, and transduction. One strategy adapts observations, \citep{church2022tactile} trains PPO policies in simulation for edge/surface following, rolling, pushing, and balancing, then uses Pix2Pix \citep{pix2pix2017} to translate real TacTip \citep{ward2018tactip} images into the simulated tactile domain for policy deployment. Another strategy simplifies the tactile representation itself. \citep{yin2025learning} uses binary normal and ternary shear signals, enabling scalable simulation and zero-shot deployment on a ReSkin-equipped \citep{bhirangi2021reskin} Allegro hand, while Text2Touch \citep{fieldtext2touch} similarly combines simulation and distillation for a TacTip-equipped dexterous hand. These approaches expose a fundamental trade-off where high-fidelity touch preserves richer interaction information but is harder to simulate, whereas abstract contact representations simplify transfer at the risk of discarding control-relevant cues. Tactile RL methods are included in Table~\ref{tab:tactile_policy_learning}.

Overall, touch contributes to RL by revealing latent interaction states, shaping exploration and rewards, and enabling deployable policies to infer physical properties represented explicitly during privileged training. Yet direct real-world tactile RL remains costly and potentially unsafe. Consequently, recent systems combine RL with pretrained representations \citep{tang2025visual,ye2026visual}, privileged distillation \citep{yin2025learning,fieldtext2touch}, structured exploration and rewards \citep{tang2025visual,fieldtext2touch}, and tactile sim-to-real adaptation \citep{church2022tactile,yin2025learning}. RL therefore increasingly serves as the adaptive policy-optimization component within a broader tactile learning pipeline rather than as unconstrained end-to-end learning from raw physical interaction.

\subsection{Diffusion and Generative Policies}

Diffusion policies have rapidly become a prominent approach to learning manipulation behaviors from demonstrations, particularly for tasks requiring coordinated and temporally extended actions \citep{chi2025diffusion, nguyen2026pixel, gu2026learning}. Their key distinction from conventional behavioral cloning lies in replacing point-wise action regression with conditional generative modeling where rather than committing to a single action for a given observation, the policy learns a distribution over plausible future action sequences. This property is particularly relevant to contact-rich manipulation, where demonstrations can be inherently multimodal. Variations in object pose, contact geometry, friction, or deformation may admit several successful interaction strategies, while averaging these behaviors through deterministic regression can produce physically ineffective actions. Diffusion policies instead preserve this diversity while generating temporally coherent action chunks, making them an attractive policy class for tactile-guided manipulation.

Given an observation history $\mathbf{o}_{1:t}$ and a future action sequence
$\mathbf{a}_{t:t+H}$, a diffusion policy represents the conditional action
distribution as
\begin{equation}
\pi_{\theta}(\mathbf{a}_{t:t+H} \mid \mathbf{o}_{1:t})
=
p_{\theta}(\mathbf{a}_{t:t+H} \mid \mathbf{o}_{1:t}),
\end{equation}
where $\mathbf{o}_{1:t}$ may comprise visual, tactile, proprioceptive, or other
task-relevant observations. Rather than directly predicting the action sequence,
diffusion policies learn to generate it through an iterative denoising process.
During training, a demonstrated action sequence $\mathbf{a}^{0}$ is corrupted at
diffusion step $k$ according to
\begin{equation}
\mathbf{a}^{k}
=
\sqrt{\bar{\alpha}_{k}}\,\mathbf{a}^{0}
+
\sqrt{1-\bar{\alpha}_{k}}\,\boldsymbol{\epsilon},
\qquad
\boldsymbol{\epsilon} \sim \mathcal{N}(\mathbf{0},\mathbf{I}),
\end{equation}
where $\bar{\alpha}_{k}$ is determined by the diffusion noise schedule. A
conditional denoising network $\boldsymbol{\epsilon}_{\theta}$ is then trained
to estimate the injected noise, commonly using the objective
\begin{equation}
\mathcal{L}_{\mathrm{DP}}(\theta)
=
\mathbb{E}_{\mathbf{a}^{0},\mathbf{o},k,\boldsymbol{\epsilon}}
\left[
\left\|
\boldsymbol{\epsilon}
-
\boldsymbol{\epsilon}_{\theta}
(\mathbf{a}^{k},k,\mathbf{o})
\right\|_{2}^{2}
\right].
\end{equation}
At inference, Gaussian noise is iteratively denoised into an action trajectory, typically executed in a receding-horizon manner \citep{mattingley2011receding}. Tactile diffusion policies therefore differ primarily in what interaction variables they generate, how touch conditions generation, and how iterative sampling is reconciled with fast contact dynamics.

\noindent Diffusion can operate at different levels of the manipulation hierarchy. PP-Tac \citep{lin2025pp} learns coordinated arm-hand trajectories for thin deformable-object manipulation from 500,000 optimized trajectories, conditioning a Transformer on proprioception and tactile deformation with an auxiliary forward-kinematics loss. TacDiffusion \citep{wu2025tacdiffusion} instead generates 6D wrench commands from force/torque history and end-effector velocity using a DDPM \citep{ho2020denoising} trained on approximately 1500 demonstrations from a behavior-tree tactile controller; the generated wrench feeds an impedance controller and transfers zero-shot across unseen insertion configurations. At a higher level of abstraction, DexterityGen (DexGen) \citep{yin2025dexteritygen} trains diffusion on approximately 10 billion transitions generated by PPO manipulation primitives to produce feasible finger-keypoint trajectories, subsequently converted to joint commands. Although DexGen uses proprioceptive rather than explicit tactile feedback, it illustrates diffusion as a reusable physical motion prior that could be further constrained by tactile interaction state.

\noindent Diffusion's iterative sampling and extended action horizons create a temporal mismatch with fast contact dynamics. Reactive Diffusion Policy (RDP) \citep{xue2025reactive} addresses this through a slow-fast architecture: a 1-2 Hz latent diffusion policy generates long-horizon action chunks from visual, tactile, and proprioceptive observations using FiLM conditioning \citep{perez2018film}, while a 20-30 Hz CNN-GRU continuously refines them using high-frequency tactile/force feedback. PP-Tac \citep{lin2025pp} similarly augments generated motions with tactile slip correction, while TacDiffusion \citep{wu2025tacdiffusion} uses a second-order filter to interface diffusion inference with a 1-kHz impedance loop. These architectures reveal a recurring hierarchy in which generative models provide slower coordinated behavior, while high-bandwidth tactile feedback stabilizes contact-level execution.

\noindent Generation quality also depends on how contact state is represented. Compact wrench or deformation signals can condition diffusion directly \citep{wu2025tacdiffusion,lin2025pp}, whereas high-dimensional touch benefits from separate representation learning. Touch in the Wild \citep{zhu2026touch} pretrains visual and tactile encoders on more than 2.6 million synchronized observations using masked tactile reconstruction and bidirectional cross-attention between CNN tactile and CLIP-ViT \citep{radford2021learning} visual features. The resulting contact-aware representation, combined with proprioception, conditions a U-Net \citep{ronneberger2015u} diffusion policy for fine-grained manipulation. Separating representation and policy learning exposes the tactile encoder to substantially greater contact diversity than task demonstrations alone, while leaving diffusion to model how these states map to future behaviors.

Overall, diffusion policies extend tactile-conditioned control beyond deterministic action regression by representing multiple feasible behaviors over extended temporal horizons. Existing methods span motion generation \citep{lin2025pp}, force-domain policies \citep{wu2025tacdiffusion}, slow-fast reactive architectures \citep{xue2025reactive}, reusable motion priors \citep{yin2025dexteritygen}, and large-scale visuo-tactile pretraining \citep{zhu2026touch}. Their central challenge is the trade-off between \emph{temporal abstraction} and \emph{contact reactivity}, longer horizons support coherent manipulation strategies, whereas rapidly changing contacts require high-frequency correction, motivating the hierarchical and hybrid architectures increasingly seen in tactile diffusion policies.



Across these learning paradigms, an orthogonal design choice is how tactile information is fused with vision and proprioception. The simplest approach concatenates independently encoded features before action prediction, enabling pretrained representations to be readily integrated into policies, as in T-DEX \citep{guzey2023dexterity}, exUMI \citep{xu2025exumi}, and deformable-object tracing \citep{zhao2026vitac}. More coupled architectures learn cross-modal interactions through fusion tokens \citep{ye2026visual} or bidirectional attention \citep{zhu2026touch}, while 3D-ViTac \citep{huang3d} and VT-Refine \citep{huang2025vtrefine} geometrically align vision and touch in a common 3D frame to explicitly ground contact within the scene. Fusion can also be temporal, Reactive Diffusion Policy \citep{xue2025reactive} separates slower multimodal action generation from high-frequency tactile correction. Thus, policy-level fusion spans feature, attention-based, geometric, and temporal integration, depending on whether touch complements visual context, spatially grounds contact, or provides rapid feedback during interaction.
\section{Tactile Foundation Models and Embodied Intelligence}
\label{sec:foundation_models}

Recent Vision-Language Models (VLMs) \citep{bordes2024introduction} and Vision-Language-Action (VLA) models \citep{kawaharazuka2025vision} are shifting robotic manipulation from task-specific policies toward general-purpose embodied models. Tactile foundation models extend this paradigm by jointly representing vision, touch, language, and robot state, aiming to transfer physical knowledge across tasks, objects, embodiments, and sensing configurations.

Formally, let
$o_t=\{I_t,T_t,p_t,l\}$,
where $I_t$, $T_t$, $p_t$, and $l$ denote vision, touch, proprioception, and language. A pretrained multimodal model learns
\begin{equation}
z_t = f_{\phi}(I_t,T_t,p_t,l),
\end{equation}
where $f_{\phi}$ is typically a large transformer \citep{achiam2023gpt, comanici2025gemini, wang2024qwen2, grattafiori2024llama} trained through self-supervised, contrastive, generative, reconstruction, or next-token objectives. The resulting representation conditions downstream action generation,
\begin{equation}
a_t \sim \pi_{\theta}(a_t,|,z_t),
\end{equation}
through imitation, diffusion, flow matching, or reinforcement learning, often while adapting only lightweight LoRA \citep{hu2024lora} layers, modality adapters, or action heads. Architecturally, these models generally combine modality-specific tokenization, transformer-based multimodal fusion, and downstream adaptation for control, reasoning, or world modeling. Recent extensions incorporate force grounding, semantic tactile pretraining, memory, predictive world models, and explicit reasoning, progressively positioning touch as a modality for embodied physical intelligence rather than solely reactive feedback.

A central challenge is grounding tactile observations in semantic spaces inherited from vision-language pretraining. Octopi-1.5 \citep{yu2025demonstrating}, built on Qwen2-VL \citep{wang2024qwen2}, projects GelSight observations through a fine-tuned CLIP encoder \citep{radford2021learning} and learns tactile attributes such as hardness and roughness through contrastive objectives. Visual Prompt Tuning \citep{jia2022visual} adapts the model for tactile description and ranking, while retrieval-augmented generation (RAG) \citep{gao2023retrieval} retrieves related tactile experiences as contextual memory. Touch thus becomes both a semantic observation and a query into prior physical experience.

OmniVTLA \citep{cheng2026omnivtla} similarly grounds touch semantically before action learning. Its dual-path tactile encoder is pretrained on ObjTac and integrated into a $\pi_0$-based policy \citep{black2024pi_0}, SigLIP \citep{zhai2023sigmoid} transfers visual semantic priors to tactile images, while a Semantic-Aligned Vision Transformer contrastively aligns touch with vision and language. The resulting embeddings encode reusable properties such as material, hardness, and roughness, illustrating a broader foundation-model principle where tactile features become more transferable when grounded in semantic or physical concepts rather than optimized solely for a single control objective.

The next step is to make these representations directly condition action generation. TLA \citep{hao2026tla} extends Qwen2-VL \citep{wang2024qwen2} toward tactile-language-action modeling by encoding sequential GelStereo observations \citep{zhang2023gelstereo} with a pretrained ViT \citep{dosovitskiy2020image}, fusing them with language tokens, and adapting the language model through LoRA \citep{hu2024lora} to autoregressively generate discretized actions. This demonstrates tactile control through a pretrained multimodal language backbone and transfer to unseen assembly geometries.

VTLA \citep{zhang2026vtla} further addresses temporal touch and the mismatch between autoregressive token prediction and continuous control. Vision-Guided Temporally Enhanced Tokens encode tactile sequences, while visual-token ordering mitigates autoregressive recency bias. After supervised next-token training, Direct Preference Optimization (DPO) \citep{rafailov2023direct} ranks sampled actions by deviation from demonstrations, providing an additional continuous-control signal. Domain-randomized demonstrations generated in Isaac Gym \citep{makoviychuk2021isaac} further support zero-shot sim-to-real insertion.

\begin{table*}[t]
\centering
\caption{Comparison of recent tactile foundation and world models for embodied robotic manipulation.}
\label{tab:tactile_foundation_models}
\renewcommand{\arraystretch}{1.25}
\footnotesize
\begin{tabular}{p{2.2cm}p{1.8cm}p{2.3cm}p{3.0cm}p{1.8cm}p{1.7cm}p{2.4cm}}
\toprule
\textbf{Model}
&
\textbf{Foundation Backbone}
&
\textbf{Tactile Encoder}
&
\textbf{Fusion}
&
\textbf{Action Generation}
&
\textbf{Primary Role}
&
\textbf{Generalization}
\\
\midrule

TLA \citep{hao2026tla}
&
Qwen2-VL
&
ViT
&
Language-conditioned tactile token fusion
&
Autoregressive
&
Foundation policy
&
Zero-shot objects
\\

Tactile-VLA \citep{huang2025tactile}
&
$\pi_0$ + Gemma
&
MLP
&
Token-level multimodal fusion + force prediction
&
Flow Matching
&
Foundation policy + CoT
&
Objects, force-sensitive tasks
\\

TaF-VLA \citep{huang2026tactile}
&
$\pi_{0.5}$
&
ViT + causal Transformer
&
Force-aligned latent representation (InfoNCE)
&
Flow Matching
&
Force-aware policy
&
Cross-sensor, objects
\\

Octopi-1.5 \citep{yu2025demonstrating}
&
Qwen2-VL
&
CLIP
&
Retrieval-Augmented Generation (RAG)
&
--
&
Tactile reasoning
&
Object retrieval
\\

VTLA \citep{zhang2026vtla}
&
Qwen2-VL
&
ViT
&
Vision-guided temporally enhanced tactile tokens
&
Autoregressive + DPO
&
Foundation policy
&
Zero-shot Sim2Real
\\

TacVLA \citep{zhang2026tacvla}
&
$\pi_{0.5}$
&
MLP
&
Contact-aware tactile gating
&
Flow Matching
&
Adaptive foundation policy
&
Visual disturbances
\\

OmniVTLA \citep{cheng2026omnivtla}
&
$\pi_0$
&
SigLIP + Semantic-Aligned ViT
&
Semantic tactile pretraining
&
Flow Matching
&
Foundation policy
&
Unseen objects
\\

FTP-1 \citep{yuan2026ftp}
&
$\pi_{0.5}$
&
Heterogeneous tactile encoder
&
Unified latent representation
&
Flow Matching
&
Generalist tactile policy
&
Tasks, sensors, embodiments
\\

VLA-Touch \citep{bi2026vla}
&
RDT + GPT-4o
&
Octopi + GelSight features
&
Hierarchical tac. planning + diffusion refinement
&
Diffusion refinement
&
Planning + control
&
Task robustness
\\

TouchWorld \citep{zhou2026touchworld}
&
Qwen3-VL + Wan2.2
&
Tactile world model + TRT
&
Hierarchical world model + residual control
&
Diffusion + residual policy
&
Embodied world model
&
Long-horizon tasks
\\

\bottomrule
\end{tabular}
\end{table*}

A complementary family extends pretrained continuous-action VLAs such as $\pi_0$ and $\pi_{0.5}$ rather than discretizing actions as language tokens. Tactile-VLA \citep{huang2025tactile}, built on $\pi_0$ \citep{black2024pi_0} and Gemma \citep{team2024gemma}, jointly encodes vision, language, proprioception, and temporally aggregated touch before a tactile-aware action expert generates continuous motion and desired contact force through Conditional Flow Matching \citep{lipman2022flow}. This allows linguistic concepts such as \textit{gently} or \textit{firmly} to directly modulate interaction through hybrid position-force control. TacVLA \citep{zhang2026tacvla} instead extends $\pi_{0.5}$ \citep{intelligence2025pi_} with contact-aware gating, activating tactile tokens only after contact to avoid interfering with pretrained visual-language reasoning during free-space motion.

However, compatibility with a VLA token space does not ensure that touch encodes the physical variables required for manipulation. This is particularly important for vision-based tactile sensors, whose image-like observations represent deformation, pressure, shear, and contact geometry rather than conventional visual appearance. The resulting question is therefore not only \emph{how} to align touch, but \emph{what latent space it should be aligned with}.

TaF-VLA \citep{huang2026tactile} addresses this through physical grounding. Its TaF-Adapter is pretrained on over 10 million synchronized tactile images, force/torque measurements, and pressure maps. Temporal tactile sequences are encoded by a ViT and causal Transformer, while force and pressure are discretized through VQ-VAE \citep{van2017neural}; InfoNCE alignment \citep{oord2018representation} then produces history-aware, force-grounded tactile embeddings without direct force regression. The frozen adapter subsequently conditions $\pi_{0.5}$ for flow-matching control across contact-rich tasks and sensors. OmniVTLA \citep{cheng2026omnivtla}, by contrast, emphasizes semantic grounding by aligning touch with vision and language. These approaches reveal complementary requirements, semantic alignment supports abstraction and transfer across objects and tasks, whereas physical alignment preserves quantities such as force and pressure required for precise interaction. General tactile foundation representations will likely need both.

A further requirement is transfer across heterogeneous tactile hardware. FTP-1 \citep{yuan2026ftp} aggregates 26 datasets spanning 21 tactile sensors to pretrain a generalist tactile policy prior on $\pi_{0.5}$. Sensor-specific observations are mapped into a unified representation shared with vision, language, and proprioception, followed by lightweight adaptation to new sensors, tasks, and embodiments. Evaluation across five robotic platforms, including unseen tactile sensors, indicates transfer beyond individual sensing configurations. Conceptually, FTP-1 moves from \emph{tactile-enhanced foundation policies}, where touch augments a pretrained VLA for particular tasks, toward \emph{foundation tactile policies}, where heterogeneous tactile experience itself contributes to the reusable prior.

Scaling such models therefore requires diversity as much as volume. Internet video can provide broad priors for generalist robot learning, but lacks aligned actions and contact information needed for tactile interaction \citep{McCarthyGeneralist25}, while tactile datasets remain fragmented across sensors, embodiments, sampling rates, and interaction protocols. ObjTac-based semantic pretraining \citep{cheng2026omnivtla}, force-grounded pretraining in TaF-VLA \citep{huang2026tactile}, and heterogeneous multi-dataset pretraining in FTP-1 \citep{yuan2026ftp} address complementary dimensions of this problem. Together, they suggest that tactile foundation models must scale across \emph{semantic, physical, and sensor diversity}, rather than sample count alone. Table~\ref{tab:tactile_foundation_models} includes main architecture and training features of tactile foundation models.

\begin{figure*}
    \centering
    \includegraphics[width=0.9\linewidth,trim={0 0 0cm 0},clip]{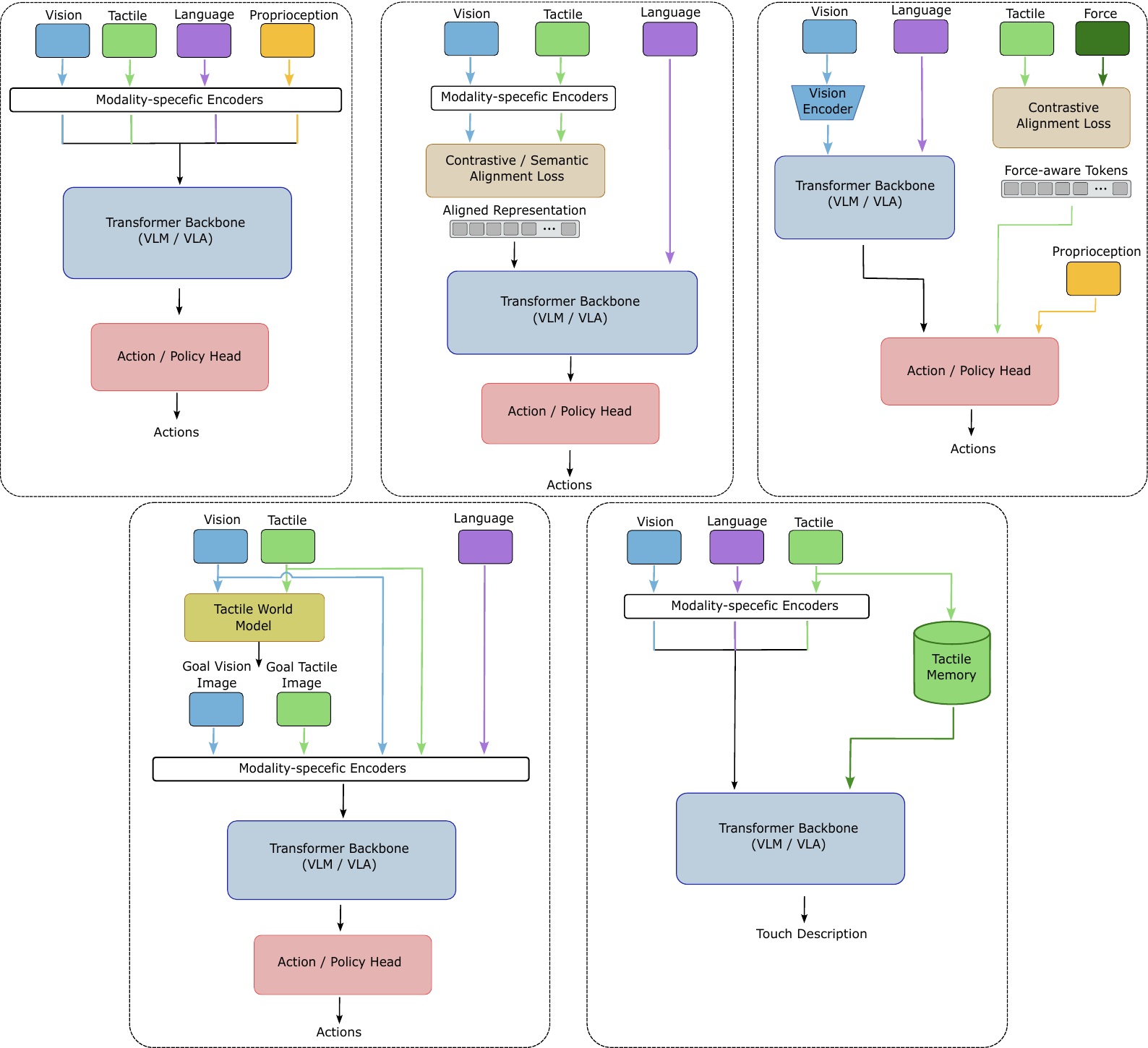}
    \caption{Representative architectures for integrating tactile information into foundation models for robotic manipulation. The paradigms include (a) direct multimodal conditioning, where tactile tokens are fused with vision, language, and proprioception within a VLM/VLA backbone; (b) semantic alignment of tactile representations with vision and language prior to policy conditioning; (c) physical grounding of tactile features through force-aligned representations; (d) predictive architectures that use tactile world models to generate future visuo-tactile interaction states for action generation; and (e) tactile-augmented reasoning, where tactile observations and retrieved interaction memories condition a VLM/VLA for semantic understanding and decision making.}
    \label{fig:vtlas}
\end{figure*}

Beyond direct policy conditioning, foundation models enable touch to support explicit reasoning and planning. Rather than mapping contact directly to action, these models can interpret what touch implies about object properties, task state, or failure and use this interpretation to guide subsequent behavior. Tactile-VLA-CoT \citep{huang2025tactile} exploits its language backbone to diagnose tactile interaction failures and generate force-aware corrections, while Octopi-1.5 \citep{yu2025demonstrating} supports semantic tactile reasoning and retrieval without an action head. VLA-Touch \citep{bi2026vla} makes this separation hierarchical, at the planning level, Octopi converts GelSight observations into language descriptions for a GPT-4o-based planner \citep{hurst2024gpt}; at the control level, Robot Diffusion Transformer actions are refined by an interpolant-based diffusion controller \citep{chen2024don} conditioned on tactile features, DINOv2 vision \citep{oquab2023dinov2}, and proprioception. These systems distinguish \emph{tactile reasoning}, which supports slower semantic interpretation and planning, from \emph{tactile control}, which requires continuous high-bandwidth physical feedback.

Most tactile VLAs, however, remain reactive, they condition actions on current observations without explicitly modeling the physical consequences of candidate actions. This distinguishes foundation \emph{policies} from embodied \emph{world models}, particularly relevant for touch because contact evolves dynamically under action. Fig.~\ref{fig:vtlas} reviews main architecture types for tactile-informed foundation models.

TouchWorld \citep{zhou2026touchworld} explicitly separates semantic planning, interaction prediction, action generation, and reactive correction. A Qwen3-VL planner \citep{bai2025qwen3} decomposes long-horizon instructions; a tactile world model adapted from Wan2.2 \citep{wan2025wan} predicts future visuo-tactile subgoals; these condition a diffusion policy generating nominal action chunks, while a higher-frequency Tactile Residual Transformer corrects execution from recent tactile and proprioceptive histories. The architecture therefore spans three timescales including semantic task reasoning, predictive interaction planning, and reactive contact regulation.

This transition changes touch from representing \emph{what contact is occurring} toward predicting \emph{what contact should occur next}. Whereas a tactile-conditioned policy models action generation from observations and language, e.g.,
\begin{equation}
p(a_{t:t+H}\mid o_{1:t},l),
\end{equation}
a tactile world model additionally captures interaction dynamics,
\begin{equation}
p(o_{t+1:t+H}\mid o_{1:t},a_{t:t+H},l),
\end{equation}
or their latent-state equivalent. Such models can predict tactile outcomes or subgoals against which candidate actions are evaluated, moving toward embodied systems that integrate perception, prediction, reasoning, and control.

Overall, tactile foundation modeling is progressing from tactile-language understanding \citep{yu2025demonstrating} and tactile-conditioned VLAs \citep{hao2026tla,zhang2026vtla,huang2025tactile,zhang2026tacvla}, toward semantically and physically grounded representations \citep{huang2026tactile,cheng2026omnivtla}, transferable generalist policies \citep{yuan2026ftp}, hierarchical reasoning and control \citep{bi2026vla}, and predictive world models \citep{zhou2026touchworld}. Yet the field remains nascent, tactile corpora are comparatively small and heterogeneous, evaluation covers limited task and embodiment diversity, and semantic reasoning, prediction, and high-frequency control remain largely modular. Achieving general-purpose tactile embodied intelligence will require models that preserve both semantic and physical interaction structure and progress from reacting to observed contact toward predicting, reasoning about, and deliberately shaping future interactions.


\section{predictive and adaptive Tactile Manipulation}
\label{sec:adaptive_manipulation}

Learned policies can acquire complex contact-rich skills, yet execution remains sensitive to changes in contact state. Tactile feedback closes this interaction loop by enabling actions, state estimates, and predictive models to adapt as contact evolves. Unlike the tactile-conditioned policies of Sec.~\ref{sec:policy_learning}, which primarily learn observation-to-action mappings, the methods considered here impose additional structure through explicit feedback regulation, state estimation, interaction prediction, online optimization, or model adaptation. Learning may remain integral, but is embedded within a broader estimation or control architecture.

\noindent \textit{Reactive tactile control.}
The most direct strategy is to modify motion in response to measured contact. FoAR \citep{he2025foar} predicts imminent contact to dynamically regulate force information within a diffusion-based visuomotor policy, suppressing force in free space and emphasizing it near contact while applying reactive position corrections. When policy inference is slower than contact dynamics, Implicit RDP \citep{chen2025implicitrdp} instead uses a slow-fast architecture, a low-frequency diffusion model generates latent action chunks, while a lightweight decoder incorporates high-frequency touch to refine execution. Related structured approaches operate without learned action policies. Tac-Man \citep{zhao2025tac} uses tactile servoing under deformation, non-slip, and contact-deviation constraints, alternating between execution and recovery; its extension \citep{zhao2026tacman} predicts object pose to anticipate trajectory corrections. For whole-arm interaction, \citep{caroleo2024proxy} combines proximity-based obstacle avoidance with higher-priority tactile force minimization. Together, these methods separate global task behavior from rapid local contact regulation, implemented either around learned policies or through explicit feedback laws.

Effective feedback also requires estimating the contact state being regulated. \citep{lloyd2024pose} represents contact pose and post-contact shear as an $\mathrm{SE}(3)$ surface-contact state, predicting its mean and uncertainty from tactile images using a Gaussian Density Network \citep{bishop1994mixture}. A Bayesian filter integrates these estimates with robot kinematics before a feedforward-feedback controller performs surface following, tracking, and pushing. Such methods combine learned tactile perception with geometrically structured estimation and control, improving interpretability while remaining dependent on estimator accuracy.

Reactive controllers correct deviations after they are observed, whereas model predictive control explicitly evaluates how candidate actions are expected to affect future interaction states. Given a current tactile or multimodal state $s_t$ and an interaction model

\begin{equation}
\begin{aligned}
a_{t:t+H}^{*}
&=
\arg\min_{a_{t:t+H}}
\sum_{k=0}^{H}
c(\hat{s}_{t+k},a_{t+k}),
\\
&\qquad
\hat{s}_{t+k+1}
=
f_{\phi}(\hat{s}_{t+k},a_{t+k}).
\end{aligned}
\end{equation}
executes the first action, incorporates the resulting tactile observation, and replans. The model $f_{\phi}$ may be specified analytically, learned from interaction data, or constructed through a combination of the two. This receding-horizon formulation makes MPC particularly suitable for tactile manipulation because model errors and unexpected contact events can be repeatedly corrected through sensory feedback.

Physically structured MPC explicitly encodes interaction mechanics. Shirai et al. \citep{shirai2023tactile} combine nonlinear trajectory optimization with quasi-static models of gripper-tool, tool-object, and object-environment contacts, using tactile state estimates to replan under disturbances. Oller et al. \citep{oller2024tactile} embed a differentiable quadratic program enforcing quasi-static equilibrium, Coulomb friction, kinematics, and tactile compliance within trajectory optimization for pushing, sliding, and pivoting. Data-Driven MPC \citep{berberich2020data, berberich2025overview} offers an alternative to explicit modeling; \citep{wang2024robust} constructs tactile dynamics directly from input-output trajectories using Hankel matrices and optimizes over these data, although its approximately linear assumptions constrain highly nonlinear contacts. These approaches trade physical structure and interpretability against assumptions that become increasingly restrictive for complex manipulation.

\noindent \textit{Learning predictive interaction models.}
Learned dynamics relax these modeling assumptions while retaining explicit online optimization. Oller et al. \citep{oller2023manipulation} predict future tactile depth maps, wrenches, and grasp poses from current touch and actions, incorporating these predictions into MPPI \citep{williams2015model}; PointNet \citep{qi2017pointnet} object embeddings condition the model across geometries, although known 3D models are required. For deformable manipulation, \citep{zhang2023visual} learns visuo-tactile latent garment dynamics through a contrastive Conditional VAE and evaluates candidate actions using complementary visual and tactile costs, with vision measuring global task progress and touch maintaining local edge contact. LeTac-MPC \citep{xu2024letac} maps tactile images into a learned latent state and embeds differentiable MPC within a 25 Hz grasp controller, optimizing gripper width online under disturbances. These methods illustrate a hybrid paradigm in which representations and dynamics are learned while action selection remains explicitly optimized.

\noindent \textit{Online adaptation of interaction dynamics.}
Even learned dynamics may fail when physical properties depart from the training distribution. RoboPack \citep{ai2024robopack} therefore infers latent object physics from sequential visuo-tactile interaction, a recurrent GNN predicts particle-level dynamics while an LSTM integrates tactile history to infer stiffness and deformability, enabling MPPI-based planning for occluded packing and non-prehensile manipulation. Adaptive contact-implicit MPC \citep{huang2024adaptive} instead combines an explicit Linear Complementarity System with a residual model updated online from prediction errors, adapting contact dynamics to unknown geometry and friction while preserving a physics prior. Here, touch identifies not only the current state but also the latent physical properties governing future interaction.

Overall, predictive and adaptive tactile control spans immediate feedback correction \citep{he2025foar,xue2025reactive,chen2025implicitrdp,zhao2025tac,zhao2026tacman,caroleo2024proxy}, probabilistic tactile servoing \citep{lloyd2024pose}, physically structured and data-driven MPC \citep{shirai2023tactile,oller2024tactile,wang2024robust}, and learned or adaptive predictive models \citep{oller2023manipulation,zhang2023visual,xu2024letac,ai2024robopack,huang2024adaptive}. Their key distinction is \emph{where learning enters the feedback loop}: perception, state estimation, dynamics prediction, model adaptation, or direct action generation. Compared with tactile-conditioned policies, these methods additionally exploit discrepancies between predicted and observed interaction during execution. As tactile world models become more expressive, this boundary will increasingly blur: MPC uses interaction models for online task-specific optimization, whereas foundation world models seek reusable predictive knowledge across broader tasks and distributions. Combining such general predictive models with MPC's constraint handling, receding-horizon adaptation, and closed-loop robustness represents a promising direction for tactile-guided manipulation.


\section{Generalization of Tactile-Conditioned Policies}

Generalization is essential for tactile-guided manipulation, yet is inherently multidimensional as a policy may transfer across objects or materials while remaining specific to a particular manipulation task, tactile sensor, or environment. This challenge is especially pronounced for touch because observations depend strongly on contact geometry, sensor morphology, deformation, and interaction dynamics, such that even small changes in the physical interface can substantially shift the tactile distribution. Robust transfer therefore requires representations and policies that preserve task-relevant physical structure while suppressing sensor- and embodiment-specific variation. Existing approaches pursue this through transferable representations \citep{higuera2025tactile}, multimodal learning \citep{tang2025visual}, simulation \citep{yin2025dexteritygen}, policy fine-tuning \citep{huang2025vtrefine}, domain adaptation \citep{jing2025reducing}, and cross-embodiment alignment \citep{zhu2026let}. However, evidence remains fragmented, with most studies demonstrating transfer along only one or two dimensions. We therefore assess generalization separately across objects, tasks, sensors, embodiments, and environments to determine how far current tactile-conditioned policies extend beyond their training conditions.

\subsection{Object and Task Generalization}

Tactile-conditioned policies must transfer when object properties or manipulation objectives differ from training. Touch can support such transfer because local interaction mechanics may remain informative despite changes in visual appearance. However, an important distinction exists between \emph{object generalization}, where the underlying skill remains similar, and the more demanding \emph{task generalization}, where physical interaction knowledge must transfer across manipulation objectives.

\noindent \textit{a) Object generalization.}
Most studies evaluate geometric transfer to unseen shapes, sizes, or contact configurations while preserving the manipulation objective. TacDiffusion \citep{wu2025tacdiffusion}, visuo-tactile insertion \citep{azulay2025visuotactile}, RoboPack \citep{ai2024robopack}, TLA \citep{hao2026tla}, Tactile-VLA \citep{huang2025tactile}, and AnySkin \citep{bhirangi2025anyskin} demonstrate varying transfer across novel geometries, where touch provides recurring physical cues such as contact alignment, force evolution, and slip. Transfer to unseen object instances is similarly demonstrated by Online Multitask Learning \citep{ye2026visual}, Touch in the Wild \citep{zhu2026touch}, PP-Tac \citep{lin2025pp}, and TactAlign \citep{wi2025tactalign}, using diverse interaction data, pretrained representations, or cross-embodiment alignment to compensate for uncertainty in object identity, configuration, or appearance.

More challenging transfer involves changes in kinematic structure or functional affordance. Tac-Man \citep{zhao2025tac} and TacMan-Turbo \citep{zhao2026tacman} manipulate unseen articulated objects without predefined kinematic models, while TEXterity \citep{bronars2024texterity}, Tactile Tool Manipulation \citep{shirai2023tactile}, and Manipulation via Membranes \citep{oller2023manipulation} accommodate variation in tools and interactions through tactile feedback or predictive models. Such results increasingly rely on shared \emph{interaction mechanics} rather than object similarity.

Nevertheless, current evaluations predominantly vary geometry or object identity in isolation; coupled changes in friction, compliance, mass distribution, surface properties, and contact dynamics remain uncommon. Existing evidence therefore demonstrates generalization mainly within structured object families rather than unrestricted out-of-distribution physical transfer.

\noindent \textit{b) Task generalization.}
Task generalization requires knowledge learned from one set of manipulation objectives to support different contact sequences, action strategies, or success criteria. This demands reusable physical representations or interaction priors rather than task-specific tactile-action correlations.

One route is task-independent tactile pretraining. Transferable Tactile Transformers (T3) \citep{zhao2024transferable} and Binding Touch to Everything \citep{yang2024binding} learn representations from diverse interactions that transfer across downstream tasks, while large-scale visuo-tactile pretraining in Touch in the Wild \citep{zhu2026touch} follows a similar principle. These methods primarily demonstrate improved downstream adaptation rather than zero-shot execution, suggesting that tactile representations currently transfer across tasks more readily than complete observation-to-action mappings.

Foundation policies offer a complementary route by integrating touch with broad semantic and action priors. Tactile-VLA \citep{huang2025tactile}, TLA \citep{hao2026tla}, and tactile-force-aligned VLAs \citep{huang2026tactile} combine pretrained vision-language-action knowledge with tactile observations that resolve task-specific physical states. Language can specify new goals, while tactile or force-grounded representations provide interaction information required to execute them, moving toward reusable semantic and physical priors rather than independent task-specific policies.

Nevertheless, task generalization must be distinguished from transfer to unseen objects, instructions, configurations, or force requirements. Most reported evaluations remain within related manipulation families or recombine familiar contact primitives such as grasping, insertion, and contact regulation; substantially fewer demonstrate fundamentally new skills absent from training. Evidence is therefore considerably stronger for object generalization and cross-task representation transfer than for unrestricted zero-shot task generalization.

\begin{table*}[t]
\centering
\caption{Taxonomy of generalization in tactile-conditioned manipulation policies.}
\label{tab:tactile_generalization}
\renewcommand{\arraystretch}{1.25}
\footnotesize
\begin{tabular}{p{2.2cm}p{3.0cm}p{3.1cm}p{3.2cm}p{1cm}p{3.0cm}}
\toprule

\textbf{Generalization Type}
&
\textbf{Transfer Scenario}
&
\textbf{Representative Strategies}
&
\textbf{Representative Methods}
&
\textbf{Current Progress}
&
\textbf{Remaining Challenges}

\\
\midrule

Object
&
Novel object geometry, size, material, friction, compliance
&
Large-scale pretraining, multimodal representations, tactile foundation models
&
TLA \citep{hao2026tla}, Tactile-VLA \citep{huang2025tactile}, TacDiffusion\citep{huang2026tactile}, Touch in the Wild \citep{zhu2026touch}
&
High
&
Large appearance and contact distribution shifts

\\

Manipulation Task
&
Different manipulation skills
&
Multitask learning, foundation models, transferable tactile representations
&
T3 \citep{zhao2024transferable}, TaF-VLA \citep{huang2026tactile}, Touch in the Wild \citep{zhu2026touch}
&
Moderate
&
Transfer across different manipulation behaviors

\\

Tactile Sensor
&
Different tactile sensing technologies
&
Sensor-invariant representations, self-supervised learning, cross-modal alignment
&
Sparsh \citep{higuera2024sparsh}, FTP-1 \citep{yuan2026ftp}, TaF-VLA \citep{huang2026tactile}
&
Moderate
&
Different sensing principles and morphologies

\\

Robot Embodiment
&
Different robot hands, grippers and manipulators
&
Cross-embodiment representation learning, latent alignment
&
TactAlign \citep{wi2025tactalign}, FTP-1 \citep{yuan2026ftp}
&
Limited
&
Different kinematics and control spaces

\\

Sim-to-Real
&
Simulation $\rightarrow$ real robot
&
Domain randomization, teacher-student learning, latent adaptation, fine-tuning
&
Reactive Diffusion Policy \citep{xue2025reactive}, PPO-based methods \citep{yin2025learning}, Touch in the Wild \citep{zhu2026touch}
&
Moderate
&
Realistic contact mechanics remain difficult to simulate

\\

Environment
&
Lighting, clutter, disturbances, object placement, contact uncertainty
&
Robust multimodal fusion, online adaptation, large-scale pretraining
&
VITAL \citep{george2024vital}
&
Moderate
&
Real-world deployment under changing environments

\\

\bottomrule
\end{tabular}
\end{table*}

\subsection{Sensor and Embodiment Generalization}

Tactile learning faces a modality-specific transfer problem because observations are intrinsically coupled to sensing hardware. Differences in transduction, resolution, compliance, and placement can produce different measurements for the same interaction \citep{gupta2025sensor,jing2025reducing}; embodiment changes additionally alter kinematics, morphology, and action spaces.

\noindent \textit{a) Cross-sensor generalization.}
Cross-sensor transfer primarily seeks representations that suppress sensor-specific variation while preserving physical interaction state. Sparsh \citep{higuera2024sparsh} and T3 \citep{zhao2024transferable} pursue this through self-supervised or heterogeneous multi-sensor pretraining, whereas UniForce \citep{chen2026uniforce} explicitly maps heterogeneous tactile observations into a force-grounded latent space using image-force mappings constrained by force equilibrium and reconstruction. This enables zero-shot transfer across GelSight, TacTip, and uSkin and cross-sensor coordination in downstream VTLA manipulation. ReactiveDiffusion \citep{xue2025reactive} evaluates control across tactile configurations, while Touch in the Wild \citep{zhu2026touch} exploits large-scale portable data spanning diverse interactions. These approaches suggest two main routes, learning invariance from sufficiently diverse tactile data or grounding heterogeneous observations in shared physical quantities such as force.

\noindent \textit{b) Cross-embodiment generalization.}
Human-to-robot transfer is currently the most developed setting. TactAlign \citep{wi2025tactalign} learns correspondence between human and robot tactile latent spaces, whereas ViTaMIn \citep{liu2025vitamin} and exUMI \citep{xu2025exumi} reduce the domain gap using portable visuo-tactile interfaces shared between human data collection and robot deployment. Visuo-tactile cortical alignment \citep{zhu2026let} similarly seeks a common human-robot latent space. Thus, embodiment mismatch can be addressed either by aligning heterogeneous sensing domains or by engineering a common observation interface.

Robot-to-robot transfer remains less established. Embeddable tactile sensing \citep{zhao2025embedding} provides a common tactile interface across structurally different hands, while FTP-1 \citep{yuan2026ftp} pretrains a shared tactile prior from heterogeneous data and adapts it across five robotic platforms. However, most approaches retain common sensing hardware or require target-embodiment adaptation; convincing zero-shot transfer across robots differing simultaneously in morphology, kinematics, sensing, and action spaces remains largely absent.

Fundamentally, sensor and embodiment shifts couple two transfer problems: \emph{observation-level correspondence}, mapping heterogeneous tactile measurements to a common physical interaction state, and \emph{action-level correspondence}, translating that state into embodiment-specific commands. Existing methods increasingly address each component, but rarely both within a unified transferable policy.

\subsection{Sim-to-Real Generalization}

Simulation enables scalable interaction, privileged supervision, and safe exploration for robot learning \citep{makoviychuk2021isaac, li2026taccel, akinola2025tacsl, tao2024maniskill3, chen2026univtac, xu2026etac}, but tactile sim-to-real is particularly challenging because touch depends directly on contact mechanics and sensor transduction. Simulators must approximate geometry, friction, deformation, force transmission, and sensor response, where even small modeling errors can alter both tactile observations and closed-loop behavior \citep{sferrazza2019ground}.

Tactile simulators therefore trade physical fidelity against computational efficiency. Physics-based methods explicitly model contact and sensor deformation, whereas graphical methods approximate sensor outputs such as depth, marker displacement, or tactile images and scale efficiently with simulators such as Isaac Gym \citep{makoviychuk2021isaac} and MuJoCo \citep{todorov2012mujoco}; reduced-order representations based on forces, contact points, or pressure provide an intermediate abstraction. Importantly, transfer does not necessarily require pixel-level realism if the policy representation suppresses sensor-specific appearance. This motivates force-grounded \citep{huang2026tactile}, geometry-based \citep{11599472}, and deliberately abstract tactile representations \citep{yin2025learning,fieldtext2touch} that seek to preserve action-relevant physical state across domains.

The sim-to-real gap can consequently arise at multiple levels: a \emph{physics gap} in contact dynamics, a \emph{sensor gap} in deformation and transduction, an \emph{observation gap} between synthetic and physical measurements, and a resulting \emph{policy gap} when these discrepancies shift closed-loop behavior. Transfer strategies target different levels of this hierarchy. Domain randomization varies geometry, sensor parameters, and initial conditions during training \citep{lin2025pp,zhang2026vtla}; privileged teacher-student learning deploys simplified binary or quantized tactile representations while exploiting richer simulated state during training \citep{yin2025learning,fieldtext2touch}; and \citep{church2022tactile} instead learns real-to-sim tactile translation, allowing policies trained on synthetic observations to operate with physical sensors.

Simulation can also complement rather than replace real data. VT-Refine \citep{huang2025vtrefine} initializes a diffusion policy from real demonstrations and uses GPU-parallel tactile simulation and reinforcement learning to discover corrective contact behaviors underrepresented in those demonstrations. Large-scale tactile pretraining \citep{higuera2024sparsh,yuan2026ftp,zhou2026touchworld} can further reduce sensitivity to exact simulated observations by learning interaction regularities across broader distributions. Overall, scalable tactile learning is unlikely to rely on a single high-fidelity simulator; more promising is to combine simulation, physical data, transferable representations, and adaptation according to which aspects of contact can be modeled reliably.


\section{Open Challenges and Future Directions}
\label{sec:challenges}


The literature reviewed in Sections~III--VII reveals a broader transition, touch is evolving from local feedback into a modality for representation, reasoning, and prediction in embodied intelligence. Yet manipulation cannot be reduced to sequence prediction because actions actively generate the observations from which contact state is inferred, and their consequences unfold through coupled interaction dynamics. Tactile observations are therefore inherently \emph{action-dependent} and contact intrinsically \emph{dynamical} \citep{johansson2009coding}. General tactile intelligence must consequently represent interaction state, predict its evolution, actively acquire missing physical information, and regulate contact across multiple timescales. This motivates a transition from \emph{tactile-augmented} policies toward \emph{tactile-native embodied intelligence}, with four directions particularly important.

\subsection{From Tactile Features to Interaction Representations}

Future tactile representations must abstract away sensor-specific form factors while retaining the physical information required for manipulation. Existing approaches span self-supervised features, semantic alignment, geometric grounding, and force-aligned latent spaces \citep{higuera2024sparsh,yang2024binding,11599472,huang2026tactile}, while cross-sensor models such as TactX and generalist policies such as FTP-1 indicate that interaction knowledge can transfer beyond individual sensing configurations \citep{park2026tactx,yuan2026ftp}. However, no representation yet provides a common interface that is simultaneously transferable, predictive, semantically meaningful, and sufficiently precise for contact control.

A promising target is an \emph{interaction-centric latent state} organized around the physics of contact rather than its sensor-specific rendering. Such a representation should preserve continuous variables-geometry, loading, deformation, slip, and compliance-while supporting semantic abstractions such as stable contact, slippage, or jamming. Bridging these levels, potentially through physical inductive biases such as friction, deformability, and history dependence, could provide a common representation for control, prediction, memory, and reasoning.

\subsection{From Tactile-Augmented Models to Hierarchical Physical Intelligence}

Foundation and generative policies expose a fundamental temporal mismatch between long-horizon reasoning and rapid contact regulation. Reactive Diffusion Policy, TacDiffusion, and TouchWorld address this through hierarchical architectures that separate slower planning or generation from faster tactile correction \citep{xue2025reactive,wu2025tacdiffusion,zhou2026touchworld}. The broader principle is that foundation models should complement rather than replace feedback control: high-level models specify intent and subgoals, intermediate models predict interactions and generate motion or force references, and low-level loops stabilize contact and reject disturbances. A central open problem is designing the bidirectional interface-what physical information should flow downward to control, and which tactile events should trigger upward replanning.

Such hierarchies could also translate abstract task goals into explicit \emph{interaction specifications}. Language describes what to achieve but rarely specifies acceptable forces, contact modes, deformation, or termination conditions. Tactile-VLA begins to connect linguistic concepts with desired force \citep{huang2025tactile}; more generally, foundation models could generate pose tolerances, contact modes, force/deformation limits, and terminal conditions that feedback controllers or MPC can enforce \citep{xu2024letac,lloyd2024pose}. This would combine semantic flexibility with precise physical control without requiring either level to subsume the other.

\subsection{From Reactive Feedback to Predictive and Active Touch}

The next major transition is from reacting to contact toward reasoning about its consequences before acting. Emerging tactile world models predict future visuo-tactile or latent states \citep{zhang2026contactworld,huang2026vitacworld,zhou2026touchworld}, but their value ultimately lies in decision making: predicting whether pressing, sliding, or regrasping will maintain stability, induce slip or jamming, exceed force limits, or deform an object. Evaluation should therefore extend beyond forecasting accuracy to whether predictions improve action selection, failure avoidance, and recovery.

Predictive models must also update when reality contradicts their forecasts. RoboPack demonstrates inference of latent physical properties from tactile history, while ReTouch adapts predictions online \citep{ai2024robopack,zhang2026retouch}. Future world models should maintain calibrated uncertainty and revise beliefs as new contact evidence arrives, particularly when prediction reliability affects safety.

Beyond prediction, tactile perception should become \emph{active}. Because touch is action-generated, robots can deliberately probe compliance, friction, grasp stability, or other uncertain properties before committing to task actions. Such epistemic actions trade immediate task progress for information gain. Integrating active touch with predictive world models would therefore shift manipulation from passive contact estimation toward purposeful physical experimentation.

\subsection{From Dataset Scaling to Lifelong Tactile Intelligence}

Scaling touch differs fundamentally from scaling vision because tactile observations vary not only with objects but also with sensor mechanics, calibration, contact pose and velocity, robot dynamics, and interaction strategy. Consequently, sheer sample count is less important than \emph{interaction diversity}. Heterogeneous pretraining in FTP-1, cross-sensor representations in TactX, and multi-sensor resources such as TacVerse support this direction \citep{yuan2026ftp,park2026tactx,wei2026tacverse}, while HapTile and FreeTacMan provide complementary mechanisms for scaling haptic-informed and robot-free demonstrations \citep{alian2026haptile,wu2026freetacman}.

Future datasets should preserve complete interaction context-synchronized actions, proprioception, vision, touch, sensor configuration, and physical measurements-and include failures, recovery, and exploratory contacts. Benchmarks should similarly evaluate coupled shifts across material, geometry, sensor, and embodiment rather than isolated transfer factors; SoftVTBench begins to incorporate deformation and physical constraints beyond nominal task success \citep{jing2026softvtbench}. Sim-to-real can be viewed as one instance of this broader transfer problem, where preserving an action-relevant physical latent state may scale better than reproducing every sensor-specific detail \citep{yang2026tactilesim2real}.

Finally, tactile intelligence must become continual. Physical interfaces are non-stationary, sensors drift or are replaced, surfaces wear, materials age, and contamination changes friction. Sensor-invariant representations and modular tactile skins provide useful foundations \citep{gupta2025sensor,bhirangi2021reskin,bhirangi2025anyskin}, but long-term autonomy requires systems that distinguish fast changes in object properties from slower changes in the sensing interface, adapt lightweight components online, and consolidate experience without catastrophic forgetting.

Ultimately, the goal extends beyond improving individual tactile policies. A tactile-native embodied agent should identify what it does not know, act to acquire that information, predict the consequences of alternative interactions, and regulate contact robustly across timescales, sensors, and embodiments. In this view, touch is not auxiliary to vision but a computational interface through which embodied models reason about and act upon the causal dynamics of the physical world.

\section{Conclusion}
\label{sec:conclusion}

This paper reviewed the transition of tactile-guided robotic manipulation from perception and reactive feedback toward learning-based physical interaction. We examined tactile representation and policy learning, foundation models, predictive and adaptive control, and generalization, highlighting how touch increasingly contributes not only to contact estimation but also to action generation, physical reasoning, and interaction prediction.

Despite rapid progress, scalable tactile intelligence remains constrained by heterogeneous sensing hardware, limited interaction data, and transfer across tasks, sensors, and embodiments. Future progress will require transferable physical representations, scalable multimodal pretraining, predictive tactile world models, and architectures that connect high-level reasoning with fast contact-level control. Together, these advances could establish touch as a core modality for general-purpose embodied intelligence, enabling robots to reason about, predict, and adapt their physical interactions in the real world.

\bibliographystyle{abbrvnat}
\bibliography{Reference}

\vfill

\end{document}